\documentclass[letterpaper]{article} 
\usepackage{aaai19}  
\usepackage{times}  
\usepackage{helvet}  
\usepackage{courier}  
\usepackage{url}  
\usepackage{graphicx}  

 \usepackage{aaai19}

\usepackage[utf8]{inputenc} 
\usepackage[T1]{fontenc}    
\usepackage{url}            
\usepackage{booktabs}       
\usepackage{amsfonts}       
\usepackage{nicefrac}       
\usepackage{microtype}      
\usepackage{graphicx}
\usepackage{caption}
\usepackage{subcaption}
\usepackage{amsmath}

\title{Efficient and Scalable Batch Bayesian Optimization Using K-Means}

\author{
  Matthew ~Groves\\
  IBM Research UK \\
  Sci-Tech Daresbury\\
  Warrington, UK. \\
  \texttt{M.J.Groves@warwick.ac.uk} \\
  \And
  Edward O.~Pyzer-Knapp\thanks{Corresponding author} \\
  IBM Research UK \\
  Sci-Tech Daresbury\\
  Warrington, UK. \\
  \texttt{epyzerk3@uk.ibm.com} \\
}

\begin{document}

\maketitle

\begin{abstract}
 We present K-Means Batch Bayesian Optimization (KMBBO), a novel batch sampling algorithm for Bayesian Optimization (BO). KMBBO uses unsupervised learning to efficiently estimate peaks of the model acquisition function. We show in empirical experiments that our method outperforms the current state-of-the-art batch allocation algorithms on a variety of test problems including tuning of algorithm hyper-parameters and a challenging drug discovery problem.  In order to accommodate the real-world problem of high dimensional data, we propose a modification to KMBBO by combining it with compressed sensing to project the optimization into a lower dimensional subspace. We demonstrate empirically that this 2-step method outperforms algorithms where no dimensionality reduction has taken place.
\end{abstract}

\section{Introduction}
\label{intro}
Bayesian optimization (BO) is a popular framework for the optimization of black-box functions, where the analytic form of the function being optimized is unknown, or too expensive to evaluate.  BO has found extensive use for the optimization of machine learning algorithms \cite{snoek2012}, and for experimental design of complex systems.

In its native form, BO is a sequential optimization procedure, since new information is required to update the posterior, and therefore the acquisition function. For many of the emerging uses of BO, this is a severe limitation since, due to the size of the optimization problem, data must be acquired in a highly parallel manner in order for the optimization to be completed in a relevant time frame. Several methods for parallelizing the BO process have been proposed, and will be reviewed in Section \ref{rw}.  It is important to point out that there are two separate, yet complementary, approaches to the parallel BO problem.  One is to minimize the strict number of function evaluations, typically achieved by a dynamically allocated batch size, and the other is to minimize the number of epochs for a given batch size.  Whilst there are situations in which both are valid, the focus of this paper is to minimize the number of epochs, since there are a number of situations in which a fixed batch size is required; for example in the screening of potential pharmaceutical compounds in which there are a pre-determined number of `slots' in which compounds can be tested. 
\section{Related work}
\label{rw}
\citeauthor{ginsbourger2007} generalize the EI to the batch setting, proposing the qEI acquisition function for batches of q points. Unfortunately, identifying the points that jointly maximize qEI is difficult, as the computational cost of evaluating the function and its derivative scales poorly with increasing q.\cite{ginsbourger2007} Several works have suggested heuristic approaches for approximating qEI, (see for example \cite{snoek2012}, \cite{chevalier2013}, \cite{wang2016}). One popular qEI-based method is the Constant Liar (CL) approach of  \citeauthor{ginsbourger2010}.\cite{ginsbourger2010}  CL is a sequential batch building method, based on iteratively adding the point that maximizes the single point acquisition function, assuming that evaluating this point will return a particular constant `lie' value, temporarily augmenting the model training set with this synthetic values and refitting the GP.

In recent work, \citeauthor{gonzalez2016} propose an alternative batching method by approximating the repulsive effect when batching.\cite{gonzalez2016} Under a Gaussian Process prior, target values of nearby points in sample space are expected to be highly correlated. Thus, when choosing a batch of samples, we may wish for the batch members to be sufficiently far apart to maximize the information gained. To do this, the authors propose the Local Penalization (LP) method that sequentially assembles batches of samples by successively penalizing the acquisition function around points previously selected, using a penalization radius based on the estimated Lipschitz constant of the acquisition function surface.

The above methods all take a greedy sequential approach to batch building, iteratively adding points to the batch that maximize a particular criterion, like the locally-penalized acquisition function. In contrast, \citeauthor{lobato2017} propose a fully parallel batch sampling technique using Thompson Sampling,\cite{thompson1933} in which the posterior is sampled to generate a `panel of experts' which are then polled in parallel as to which data point should be acquired.\cite{lobato2017}

\citeauthor{ppes} suggest Parallel Predictive Entropy Search (PPES), a non-greedy batch sampling approach aiming to maximize the expected information gain from sampling the chosen batch in terms of the expected reduction in differential entropy of the predictive distribution of the global maximizer given the sampling data.\cite{ppes}

\citeauthor{b3o} propose a novel batch selection method called Budgeted Batch Bayesian Optimization (B3O), which aims to build sample batches containing peaks of the acquisition function.\cite{b3o} To find these peaks, whilst avoiding costly optimization routines, the authors propose a generalized slice sampling procedure. Slice sampling preferentially accepts samples from high density regions of the acquisition function surface, allowing peaks to be reliably estimated even with modest numbers of samples. Peak picking is then done using an Infinite Gaussian Mixture Model (IGMM) \cite{rasmussen2000}. \citeauthor{b3o} show empirically that B3O performs well on a variety of test functions and common BO applications, such as hyperparameter tuning. However, the inability of B3O to allow fixed batch sizes is a potential limitation, as real-world applications for batch BO can have an effective constraint on possible batch size, for example the number of available compute nodes (simulation), number of different molecules that a robotic assay can test simultaneously (drug discovery), or quantity of samples that can fit in a furnace (alloy hardening). Under-utilizing the available resources with smaller batch sizes costs information that could be gained at little additional cost, whereas choosing to allocate too many samples to a batch may be impossible.

\section{Proposed Method}
\label{pm}

In the BO formalism, the target function is not directly optimized.  In its place, an acquisition function is constructed using a probabilistic model based upon previously determined values for the function $f$.  Typically this model is a Gaussian process (GP),\cite{rasmussen2006gaussian} although other models including neural networks have been used.\cite{snoek2015scalable}  

There are many different versions of the acquisition function, depending upon the type of optimization task which is being performed, but the most commonly used is expected improvement, EI,\cite{mockus_bayesian_1974} which is determined as follows:
\begin{equation}
EI(x) = \mu(x) - f^{*}\Phi(\gamma) + \sigma(x) \phi(\gamma)
\label{eq:ei}
\end{equation}
where $\Phi$ denotes the CDF (cumulative distribution function) of the standard normal distribution, $\phi$ denotes the PDF (probability density function) of the standard normal distribution, and $\gamma$ denotes the improvement, which can be expressed as:
\begin{equation}
\gamma(x)  = \frac{\mu(x) - f^{*}}{\sigma^{2}(x)}
\label{eq:improvement_max}
\end{equation}
where $f^{*}$ is the best target value observed so far, $\mu(x)$ is the predicted mean and $\sigma^{2}(x)$ is the corresponding variance. 

At its core, this procedure is inherently serial, as it is based upon the updating of a probablistic model, and thus limited by data acquisition.  Our contribution is twofold, firstly we propose a novel parallel (or batch) Bayesian optimization procedure based upon K-means, K-means Batch Bayesian Optimization (KMBBO), and secondly we propose a modification for the use of this method with very high-dimensional data using a dimensionality reduction step based upon compressed sensing. 

The central aim of KMBBO is to efficiently select a batch of high quality points to evaluate, i.e, during each sampling epoch, we would like our batch to contain points from high-density regions of the acquisition function. However, modeling the landscape of the acquisition function directly is generally intractable, except in very low dimensions. In order to approximately learn the locations of peaks, we fit a K-Means clustering model to the collection of points in our sample space, chosen using slice sampling. Slice sampling draws samples uniformly from the volume under the acquisition function, and so will preferentially select samples from regions where the acquisition function value is highest.

K-Means \cite{macqueen1967} is one of the simplest and most commonly used clustering methods. Given a set of points $X$, and number of clusters $k$, the K-Means method will attempt to find a partition $P(X) = \{ X_{1},...,X_{k} \}$ clustering the members of $X$ in order to minimize the within-cluster sum of squares distance between cluster members and the cluster centroid, i.e:
\begin{equation}
\underset{P(X)}{argmax} \sum_{i=1}^{k} \sum_{x \in X_{i}} ||x - \mu_{i}||^{2}
\end{equation}
where $\mu_{i}$ represents the centroid of cluster $i$. 
Thus, KMBBO allows the user to specify the batch size directly as the number of clusters for the K-Means method.
\begin{table}
\centering
\caption{K-Means Batch Bayesian Optimization (KMBBO)}
\resizebox{.5\textwidth}{!}{
  \begin{tabular}{lll}
  \hline
Input: &\multicolumn{2}{l}{Sampling domain $\mathcal{X}$, Initial samples $\mathcal{D}_{0}$, Batch size $k$,}\\
&\multicolumn{2}{l}{epochs $N$, slice samples $n_{s}$}\\
&\multicolumn{2}{l}{Batch size $k$,epochs $N$, slice samples $n_{s}$}\\
For t = 1 to $N$:  &&\\
&\multicolumn{2}{l}{1. Fit GP model to training data $\mathcal{D}_{t-1}$}\\
&2. Collect slice samples: & $\{s_{1},...,s_{n_{s}} \} = BGSS(\mathcal{X})$\\
&\multicolumn{2}{l}{3. Fit K-Means model to obtain centroids $\mu_{i}$}\\
& 4. Sample centroids: & $\{y \}_{t} = \{f(\mu_{1}),...,f(\mu_{k}) \}$\\
& \multicolumn{2}{l}{5. Add newly observed values to dataset:}\\
&& $\mathcal{D}_{t} = \mathcal{D}_{t-1} \cup \{y\}_{t}$\\
End for &&\\
Return $\mathcal{D}_{N}$ &&\\
\hline
  \end{tabular}
}
\end{table}

To collect its slice samples, KMBBO utilizes the batch generalized slice sampling (BGSS) method described in \cite{b3o} where the joint density is defined as
\begin{equation}
p(u,u)=
\begin{cases}
\frac{1}{z},  \text{ if } \alpha_{min} < u < \alpha(x)\\
0,  \text{ otherwise }
\end{cases}  
\label{eq:bgss}
\end{equation}

where $z = \int \!\alpha(x) dx$ and $\alpha_{min}$ is obtained through minimization using a non-convex global optimizer, thus not requiring the function to be non-negative, or a proper distribution. However, like standard slice sampling, BGSS scales poorly with the dimensionality of the sampling domain \cite{neal2003}, making it impractical for use in high-dimensional settings. To address this we add a dimensionality reduction step based upon compressed sensing.  Our use of the compressed sensing methodology is based upon the observation that most high-dimensional data follows a sparse encoding and thus is compressible.  In the compressed sensing scheme, the aim is to reconstruct a signal using the smallest number of observations (which are linear functions of the components of the signal) possible. This is achieved by solving the basis pursuit problem, where we search for the sparsest matrix $A$ which can reconstruct the full matrix $B$:
\begin{equation}
    min\|A\|_{1} \text{ subject to } (PAP^{T})_{ij} = B_{ij} \text{   } \forall i, j \in W
\end{equation}
where $P$ is the change-of-basis matrix and $W$ is a set of randomly measured entries in matrix B.

We apply this method to the original feature space of a high dimensional problem, but instead of using the sparse solution to reconstruct the original function, we instead use it as a compressed basis in which to perform the BO sampling. 

The upper bound on the lossless dimensionality reduction which can be achieved using compressed sensing is thus equivalent to the number of samples which are required for compressed sensing to perfectly recover $B$ from $A$, which has been shown to scale as follows: \cite{candes2008introduction}:
\begin{equation}
    M \propto \mu^{2}S\text{log}(N)^{2}
    \label{eq:scaling_CS}
\end{equation}
where $N$ is the original number of features, $S$ is the number of non-zero elements and $\mu$ is the incoherency, which in general ranges from $1$ to $\sqrt{N}$.

\begin{table}
\centering
\caption{KMBBO with compressed sensing (CS-KMBBO)}
\resizebox{.5\textwidth}{!}{
  \begin{tabular}{lll}
  \hline
Input: & \multicolumn{2}{l}{Domain $\mathcal{X}$, compression error tolerance $\epsilon$,}\\ 
& \multicolumn{2}{l}{Batch size k,\# epochs N, \#slice samples $s$}\\
1. Draw $n_{comp}$ samples from $\mathcal{X}$: &\multicolumn{2}{l}{$S = \{ s_{1},...,s_{n_comp}\}$ } \\
2. Compress domain using TwIsT: & \multicolumn{2}{l}{$\mathcal{X}_{cs} = Compress(\mathcal{X},S, \epsilon)$ }\\
3. Run KMBBO: & \multicolumn{2}{l}{$\mathcal{D}_{N} = KMBBO(\mathcal{X}_{cs}, k,N,s)$}\\
4. Decompress $\mathcal{D}_{N}$&&\\
\hline
  \end{tabular}
}
\end{table}

Whilst some other methods, such as REMBO,\cite{wang2013bayesian} have used a compressive scheme, the exact dimensionality of this compression was left as a parameter to tune.  In CS-KMBBO, we instead use the Two-step Iterative Shrinkage/Thresholding (TwIsT)\cite{bioucas2007new} optimization technique - a variant of the popular Iterative Shrinking Thresholding algorithm (IsT) which is more robust to ill defined measurements - to determine the optimal dimensionality of the compression step.  Whilst it is possible in some discrete problems, such as the drug discovery challenge tackled within this paper, to know the entire space of inputs, we recognize that this is not always the case.  Thus, we sample 1,000 data points to perform the TwIST-based dimensionality optimization procedure to create a process which is transferable between discrete and continuous spaces.      

\section{Experiments}

\subsection{Comparison to Existing Methods}

For this study we compare the performance of KMBBO to a range of currently used parallel BO methods, the details of which have been described in Section \ref{rw}, using the Expected Improvement acquisition function, which has been shown to have strong theoretical guarantees \cite{vazquez_convergence_2010}and empirical effectiveness \cite{snoek2012}. In addition to Naieve qEI, the most basic parallel sampling method, we compare to Thompson sampling, Constant Liar (mean), Local Penalization, a batch predictive entropy search model to represent a non-greedy search strategy, and the dynamic batch method B3O.  We investigate two metrics for success:
\begin{enumerate}
\item The convergence of the search to the global minimum (where known) as a function of the number of epochs
\item The robustness of the search, as demonstrated through sampling 100 repeat runs of the sampling experiment.
\end{enumerate}

For this study, a batch size of 8 was arbitrarily chosen. 
Throughout the study the Bayesian model was provided through the use of a Gaussian process, which was created using a squared-exponential kernel with automatic relevance determination (ARD) as implemented in the Scikit-Learn library \cite{scikit-learn},
\begin{equation}
k_{SE} = \sigma_{f} exp \frac{-1}{2} \sum^{D}_{d=1}\frac{-(x_{d}-x^{*}_{d})^{2}}{2l_{d}^{2}})
\label{eq:se_kernel}
\end{equation}
seeded with 10 randomly selected data points. The GP's hyperparameters were optimized using gradient descent on the marginal likelihood. Finally, both B3O and KMBBO selected 200 slice samples when generating each batch to maintain consistency with \cite{b3o}.

\subsection{Optimization Tasks}
\subsubsection{Synthetic Functions}

We test the ability of KMBBO to find the global extremes of three synthetic functions commonly used for benchmarking machine learning algorithms: Branin-Hoo (2D), Camelback-6 (2D),  and Hartmann (6D) as described on the Virtual Library of Simulation Experiments test function database \cite{simulationlib}.

\subsubsection{SVM}
A common use for Bayesian optimization is for the tuning of hyperparmeters for machine learning models. In order to test the effectiveness of KMBBO for this task, we use it to determine optimal hyperparameters for a support vector machine for the Abalone regression task.\cite{nash_population_1994}  In this context we tune three hyperparameters:  $C$ (regularization parameter), $\epsilon$ (insensitive loss) for regression and $\gamma$ (RBF kernel function). The loss function is the root mean squared error of the prediction. 

\subsubsection{Drug Discovery}
This is a task taken to illustrate the utility of this procedure for lead identification in drug discovery -  where rapid identification of desirable compounds at low cost is essential.  The target for maximization is the PEC50; a value which describes the potency of the drug. The data was taken from hits from Plasmodium falciparum (P. falciparum) whole cell screening originates from the GlaxoSmithKline Tres Cantos Antimalarial Set (TCAMS), Novartis-GNF Malaria Box Data  set and St. Jude Children's Research Hospital’s Dataset (EC50 in $\mu$M against P. falciparum 3D7) as released through the Medicines for Malaria Venture website \cite{noauthor_malaria_nodate}. Each molecule was described using MAACS keys \cite{maccs}- a common cheminformatics descriptor, generated using the RDKit software \cite{landrum_rdkit:_nodate} resulting in a 167 dimensional optimization problem. 
\section{Results}
\begin{figure}[hthp] 
   \includegraphics[width=\linewidth]{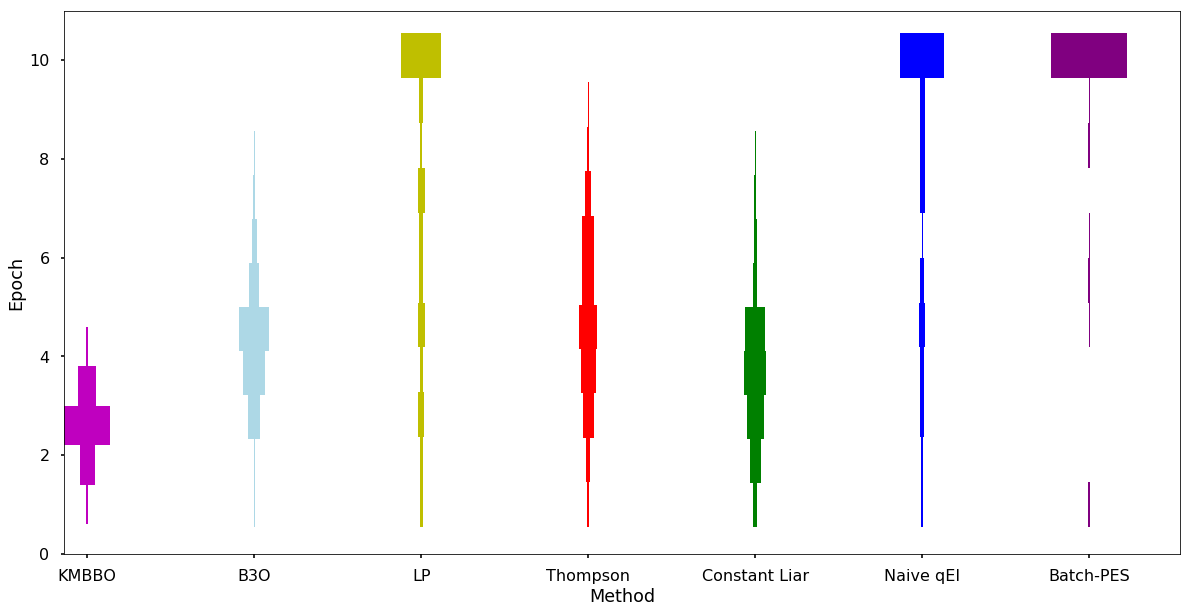}
\caption{The distribution of the `first encounter time', i.e. when the global optimium is first located for the Branin-Hoo function.  Statistics are generated from 100 repeats of the experiment.}
 \label{fig:1} 
\end{figure}

\label{exp}
\subsection{Synthetic Functions}
For the Branin-Hoo, we observe that both the Constant Liar and KMBBO methods are able to approach the minimum quickly, achieving low regret after only a few sampling epochs, with both B3O and Thompson sampling also reliably reaching finding the optimum before 8 sampling epochs. LP, however, performs poorly, achieving similar regret to Naieve qEI, with many iterations of each method failing to discover the minimum after 10 epochs. The performance of LP relies heavily on the quality of the Lipschitz constant estimate, which is calculated over the entire sampling domain. For the Branin-Hoo function, this is dominated by the quartic term away from the function minima, leading to a Lipschitz constant estimate poorly suited to the region around the optimum. Figure \ref{fig:1} shows the `first encounter time' of the global optimum for each method. We see that, even though the initial reduction in regret between KMBBO and Constant Liar is similar, KMBBO is able to locate the optimal value earlier and more consistently than the other methods.  All functions perform well for the Camelback task, although we observe that KMBBO converges to the true minimum faster than the other methods. 
\begin{figure}[hthp] 
   \includegraphics[width=\linewidth]{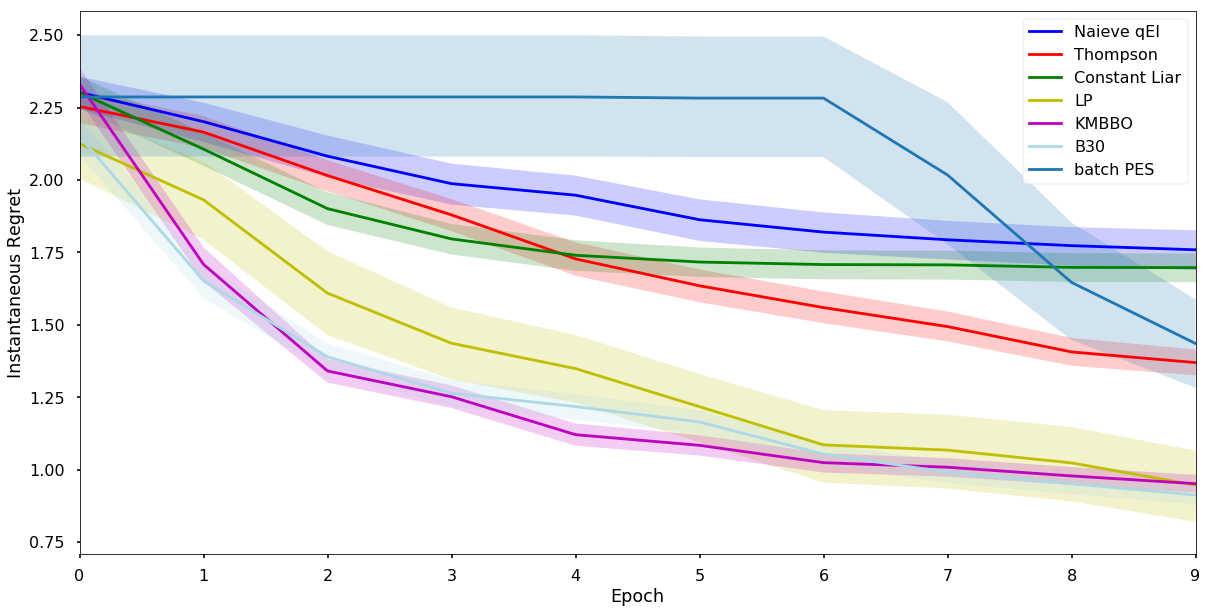}
\caption{The optimization performance for the 6 dimensional Hartmann function.  Statistics are generated from 100 repeats of the experiment, and confidence intervals are calculated to 1 sigma.}
 \label{fig:2} 
\end{figure}
The 6 dimensional Hartmann function is a more challenging optimization problem. We observe in Figure \ref{fig:2} that after 10 epochs are methods have still not yet managed to identify the global optimum. LP, B3O and KMBBO all performed well, achieving similar average regret values, but B3O and KMBBO performed more consistently, with lower variance on the regret obtained.

\subsection{Tuning of Hyperparameters}
\begin{figure}[h!] 
   \centering
   \includegraphics[width=\linewidth]{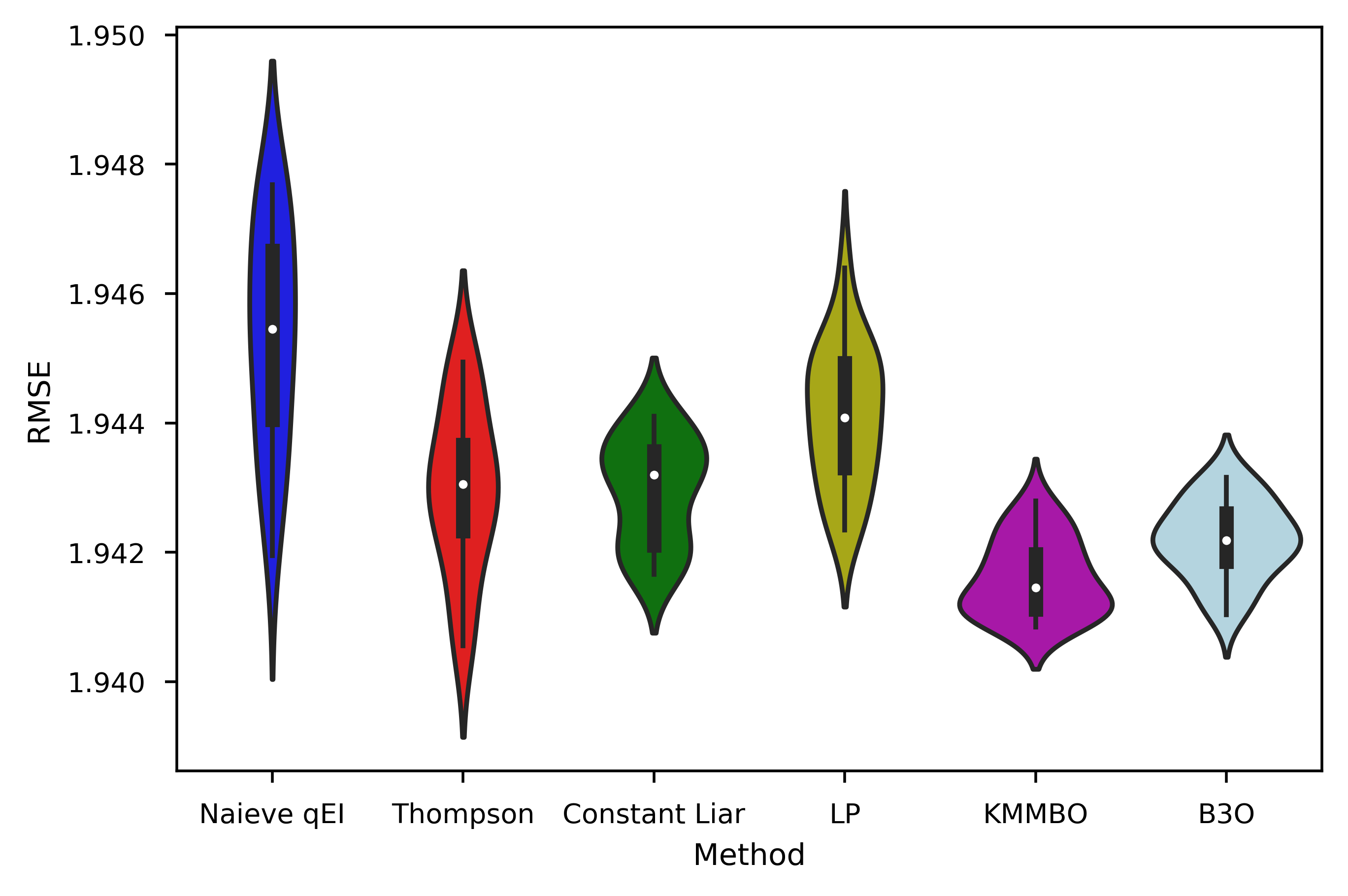}
   \caption{}
\caption{Optimization performance for the tuning of the hyperparameters of an SVM, as displayed through the RMSD of the SVM for the abalone problem with respect to the number of epochs.  Statistics are generated from 100 repeats of the experiment, and confidence intervals are bootstrapped to 1 sigma.  Note that the batch-PES methodology is excluding from this plot, as its large variance over runs made interpretation of the performance other methodologies impossible.  The results for this method can be seen in Table \ref{results_table}}
 \label{fig:svm}
\end{figure}

KMBBO displays the best performance on the SVM hyper-parameter tuning task, shown in Figure \ref{fig:svm}. With a low dimensional sampling space, the slice sampling method used by B3O and KMBBO performs particularly well at approximating high density regions of the acquisition function. Indeed, the violin plot in \ref{fig:svm} shows that, not only are KMBBO and B3O the best performers at minimizing RMSE, they also perform most consistently, with smallest error variance.

\begin{table*}[h]
\centering
\caption{Final performance after 10 sampling epochs for each method on each of the test problem cases over 100 repetitions. Best performance in each case is shown in bold. For the Malaria task, an X indicates that the method was not run, due to computational intractability or algorithmic instability.}
\resizebox{\textwidth}{!}{
\begin{tabular}{lcccccccccc}\toprule
&\multicolumn{8}{c}{\textbf{Task}} \\
\cmidrule(r){2-11}
&\multicolumn{2}{c}{\textbf{Branin-Hoo}} & \multicolumn{2}{c}{\textbf{Camelback-6}}&\multicolumn{2}{c}{\textbf{Hartman}}&\multicolumn{2}{c}{\textbf{SVM}}&\multicolumn{2}{c}{\textbf{Malaria}}\\
\midrule
\textbf{Method} & Regret & Std.Dev& Regret & Std.Dev& Regret & Std.Dev& RMSE & Std.Dev& Regret & Std.Dev\\
\midrule
Naieve qEI&0.803&1.47&\textbf{0.0276}&0.0974&1.74&0.700&1.9453&0.00170&3.1185 &0.8392\\
Thompson&0.00619&0.00186&0.0727&0.179&1.33&0.438&1.9430&0.00125&3.0000&1.0888\\
Constant Liar&0.00584&0.00129&0.0778&0.207&1.70&0.511&1.9430&0.000802&X&X\\
LP&0.637&1.28&0.0292&0.0947&0.916&0.673&1.9441&0.00105&X&X\\
KMBBO&\textbf{0.00523}&0.000488&0.0354&0.0616&0.922&0.311&1.9416&0.000577&\textbf{2.3802}&1.4003\\
B3O&0.00591&0.00170&0.130&0.338&\textbf{0.882}&0.320&1.9422&0.000580&X&X\\
Batch PES&0.5486&0.3682&0.1619&0.0967&1.4257&0.4736&\textbf{1.9406}&0.7322&3.2626&0.7322\\
\bottomrule
\end{tabular}%
}%
\label{results_table}
\end{table*}

\subsection{Drug Discovery}

The high dimensional nature of the drug discovery task presented significant challenges to several of the benchmark methods. In 167 dimensions, the slice sampling method used by B3O is unable to produce any reasonable approximation of the acquisition function surface with the original sampling budget of 200 and we found the substantial increase in samples required lead to prohibitively long running times. The LP method was hamstrung by the computational cost of approximating the Lipschitz constant in this high dimensional space, Furthermore, the Constant Liar methodology is reliant upon a high quality model, and thus is very sensitive to hyperparameter selection, and the addition of reasonable quality psuedo-inputs.  Unfortunately, during our testing of this method for the drug discovery problem, a large number of runs failed due to a failure for the GP model to converge during the fitting task, and thus it is excluded from the results.   

\begin{figure}[h!] 
   \centering
   \includegraphics[width=1.0\linewidth]{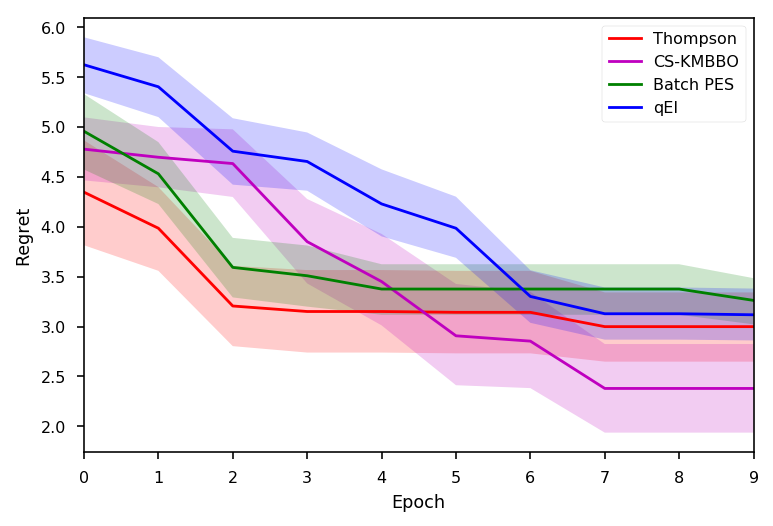} 
\caption{Optimization performance for the Malaria drug discovery problem, as displayed through instantaneous regret with respect to the number of epochs.  Statistics are generated from 10 repeats of the experiment, and confidence intervals are bootstrapped to 1 sigma.}
 \label{fig:malaria}
\end{figure}

Of the remaining methods, Thompson sampling, qEI, batch-PES and CS-KMBBO are able to be used for this task.  Figure \ref{fig:malaria} shows that KMBBO displays strong performance, reaching low regret after 10 sampling epochs- having sampled only 90 out of a potential circa 19,000 candidate molecules. Thompson sampling, qEI and Batch-PES display similar behaviors, discovering a local maximum on the potency landscape, but neither are able to discover molecules with as low regret as KMBBO.  It is worth noting that due to the discrete nature of the search space, here regret is not a continuous function, for example a regret of 2 places you within the top 0.7\% of values for the task. 

\subsection{Rankings}

One way to measure the robustness of a search method is to compare the rankings of the search method of the whole range of tasks performed in this study. Since raw rankings can be misleading (a close second ranks the same as a search in which the gap between methods was much wider) we instead use a normalized ranking, $Z$, proposed in \cite{jasrasaria_dynamic_2018}:
\begin{equation}
Z = \frac{s - s\prime}{s_{max} - s_{min}}
\label{eq:norm_rank}
\end{equation}

where $s$ represents the result of a particular strategy, $s\prime$ the result of the best strategy, and $s_{max} - s_{min}$ represent the range of results encountered in the study. 
This results in a score bounded $(0,1)$ where 0 represents a perfect performance across tasks. 

We calculate $Z$ for both the performance of the optimization task, as measured by regret or RMSE where appropriate, and the variance of the task as measured over multiple runs. 

\begin{figure}[h!] 
   \centering
   \includegraphics[width=1.0\linewidth]{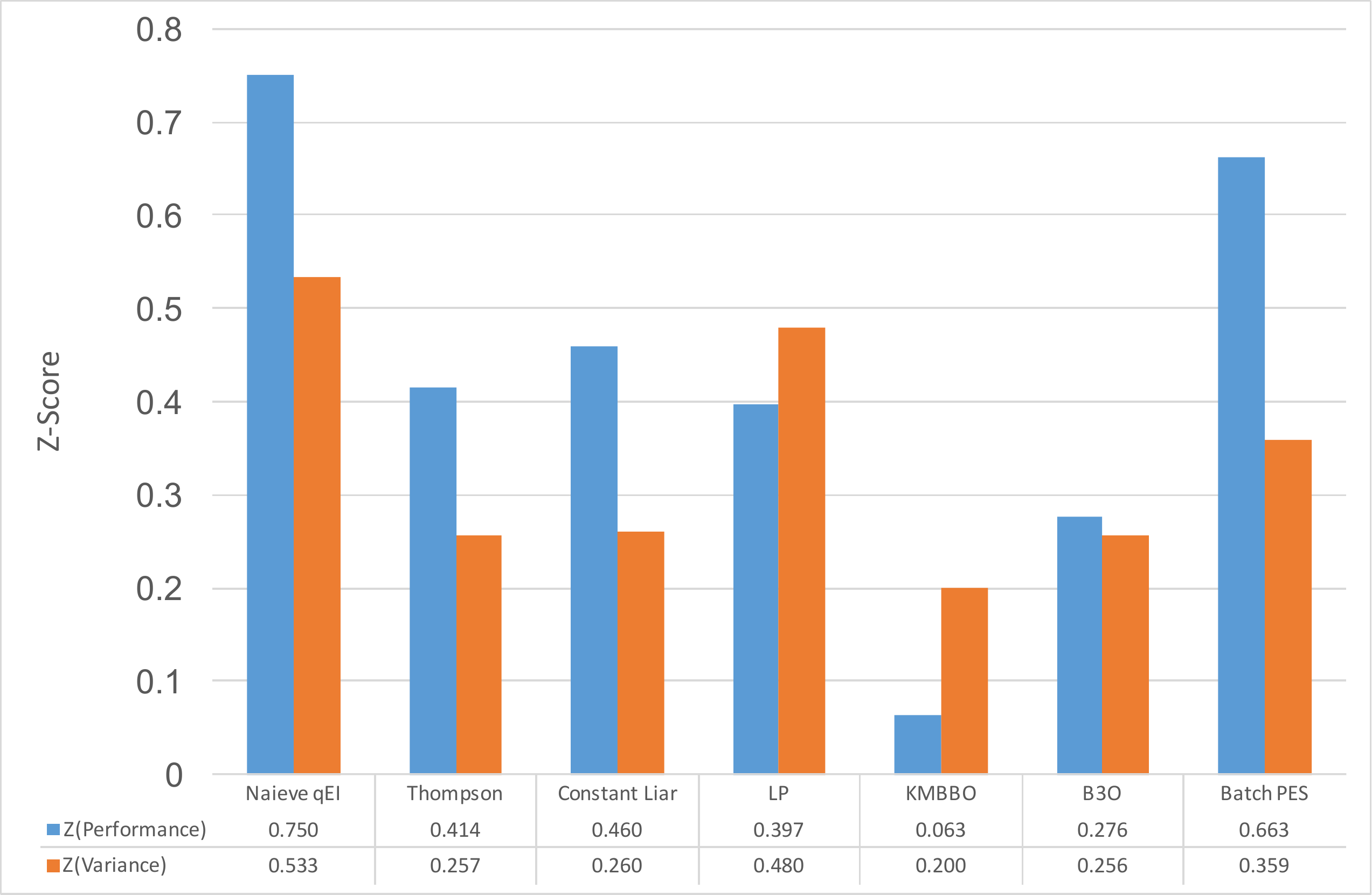} 
\caption{Z score calculated for each of the parallel optimization strategies investigated in this study.}
 \label{fig:zscore}
\end{figure}

It can easily be seen from Figure \ref{fig:zscore} that KMBBO achieves a significantly better Z score than any other method for pure optimization performance, and also the best Z score, albeit by a smaller margin, than any other method for variance. This demonstrates both the class leading nature of KMBBO and also its strong reproducibility; a property which is key for Bayesian optimization, where each data point is expensive to acquire and thus reliability of a methodology is strongly desired.  

\subsection{Computational Cost}
\begin{figure}[hb]
    \centering
    \includegraphics[width=\linewidth]{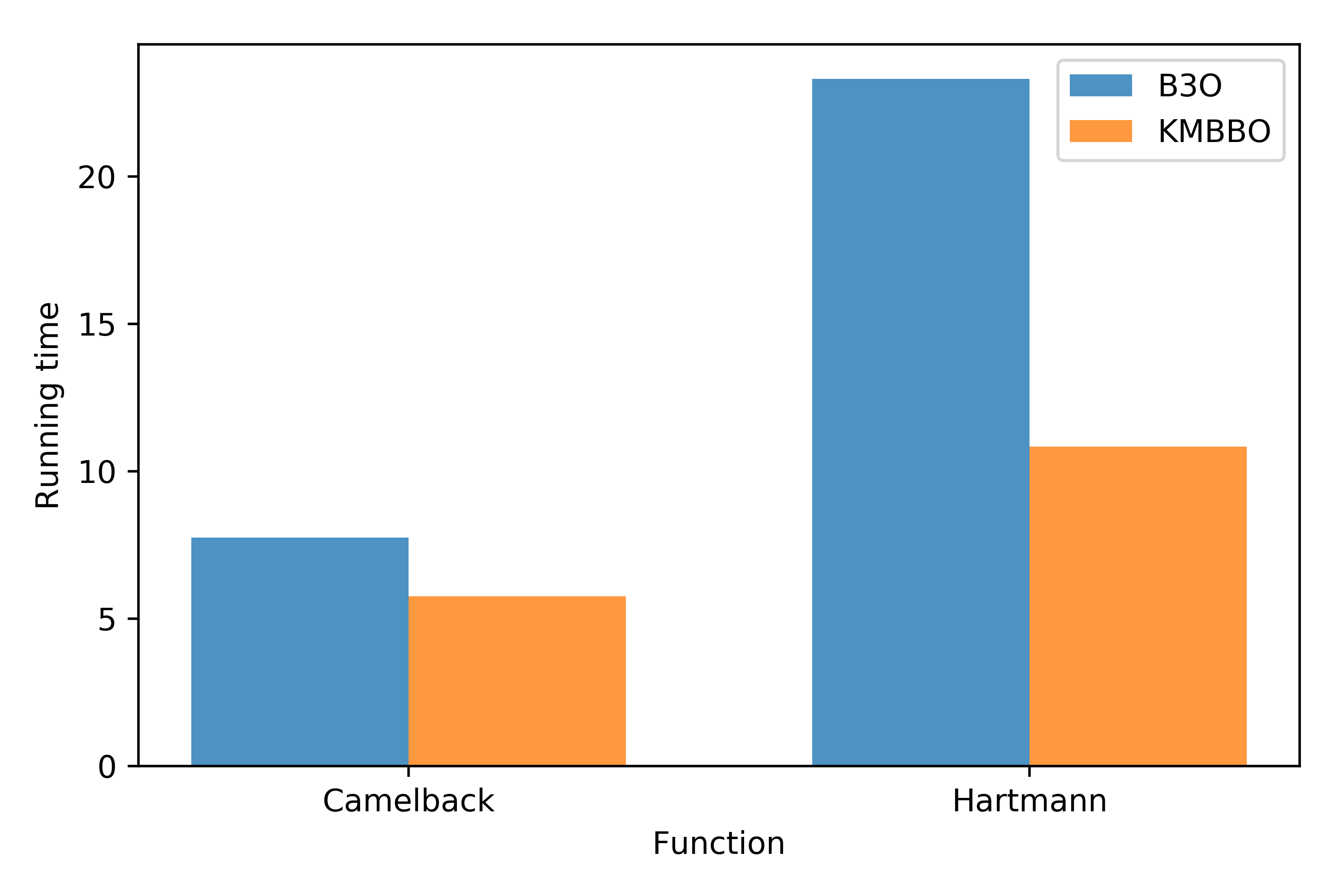}
    \caption{Runtimes of KMBBO and B30 in 2 and 6D.  Runtime is calculated as seconds per sampling epoch.}
    \label{fig:runtimes}
\end{figure}

 We have analyzed the complexity of the rate-limiting step for each of the methods used in this work, and performed additional empirical experiments looking at real-world running times . The poor dimensionality scaling of slice sampling ($O(2^{d})$) is common to the B3O method, and worse than the scaling of the LP method ($O(d^{3})$). We address this in  CS-KMBBO through the incorporation of compressed sensing for dimensionality reduction. Even when compression is not required, our empirical timings, shown in Figure \ref{fig:runtimes}, indicate that the runtime per sampling epoch for KMBBO is generally significantly smaller than for B3O, which we attribute to the simplicity and scalability of the K-Means algorithm compared to the IGMM used in B3O. However, it is worth keeping in mind that in the Bayesian Optimization framework, it is generally assumed that obtaining ground truth values by sampling the black-box function $f$ is substantially more expensive (in time/ computational cost) than the calculation of the sampling batch, which somewhat mitigates concerns about the computational cost of the sampling methodology as an expensive, yet efficient, sampling scheme will have less real-world cost than an inefficient, yet fast, alternative.
 
\subsection{Algorithmic Insight}
In this section we discuss the different characteristics of the sampling methods through analyzing their sample selections for an easy to visualize 1 dimensional optimization problem.
\begin{figure}[t] 
   \centering
   \includegraphics[height=5cm]{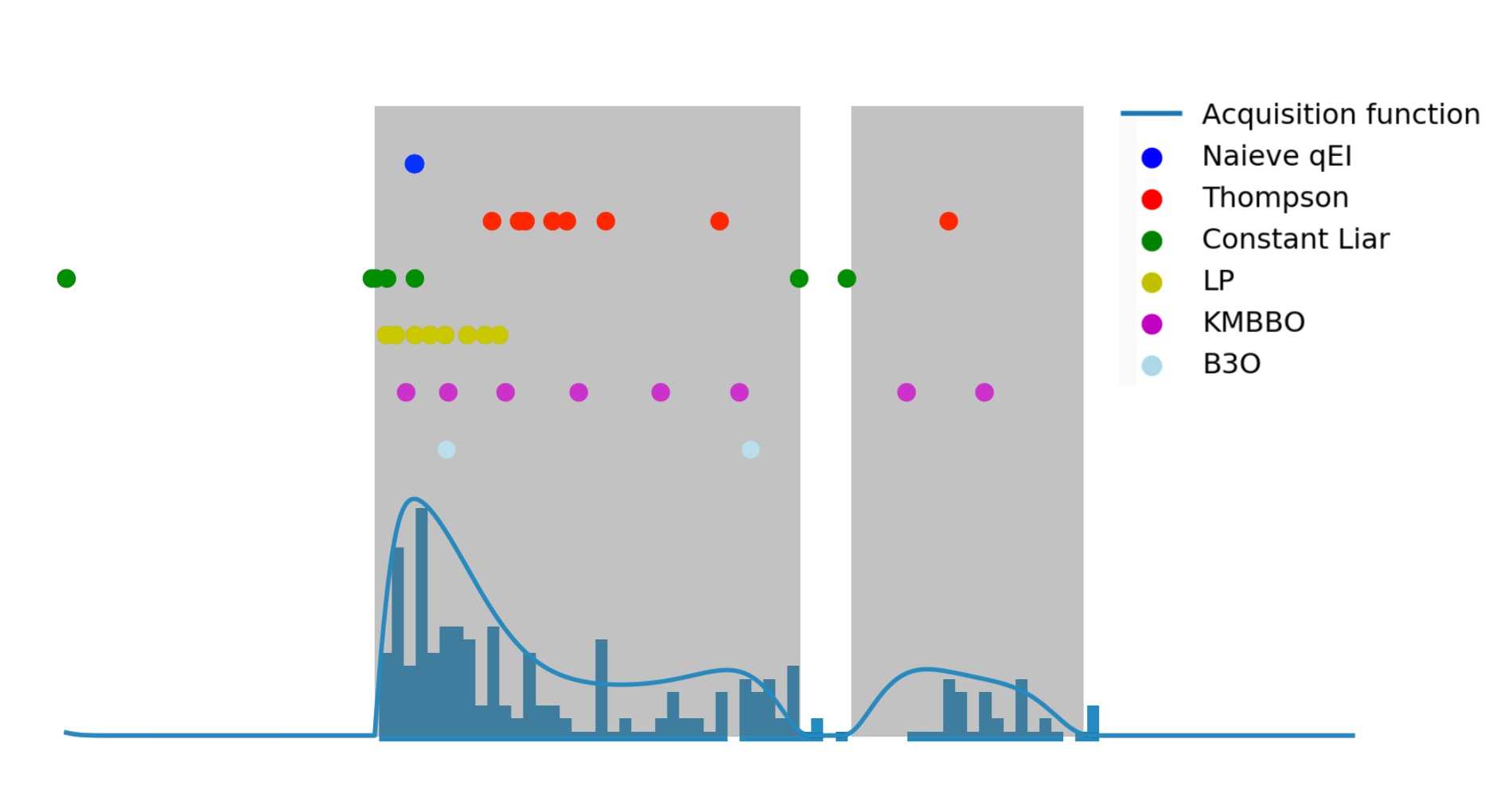} 
\caption{Points selected to form the next batch for each sampling method when minimizing $f(x) = -xsin(x)$, given 5 initial random points. The activation function is shown in blue, with non-zero regions shaded. The blue histogram shows the samples taken by BGSS.}
 \label{fig:sp}
\end{figure}
Figure \ref{fig:sp} shows the activation function curve and subsequent samples chosen by each of the sampling algorithms while minimizing the function $f(x) = -xsin(x)$, after 5 randomly chosen initial samples. This gives some visual insight into the behavior of each of the methods. We observe that all of the methods are able to identify the main peak of the acquisition function and allocate at least one sample nearby. Naieve qEI simply chooses the q points from the sample space closest to this peak, leading to highly local sampling, and insufficient exploration of other areas of density in the acquisition function. LP also does a good job of identifying the main acquisition function peak, and the local penalization factor ensures somewhat more exploration than with the Naieve qEI method. However, this still seems insufficient to cause exploration of other areas of density in the acquisition function. In contrast, Constant Liar is susceptible to over exploration and selects several low quality points. We posit that this is due to the assumption that the true value for each sample added to the batch is represented by the mean value of the GP prediction.  Since the violation of this assumption can lead to large movements in the GP posterior, this can cause erratic behavior, and lead to these poor selections. In our toy example, B3O successfully identifies two of the acquisition function peaks, but does not represent the third. The IGMM used by B3O seems to be sensitive to the number of slice samples provided, as experiments with different numbers of slice samples lead to substantial variations in the number and location of the points chosen. 

KMBBO is able to achieve a good balance between exploration and exploitation, with all three maxima in the acquisition function represented, with the remaining samples well distributed over the non-zero areas of the curve. When the number of local optima of the acquisition function is lower than the batch size, the quadratic penalization for within cluster distance used by K-Means ensures that the remaining cluster centroids will spread out over the set of slice sample values.

\section{Summary}
\label{sum}
We propose a novel batch sampling algorithm for Bayesian optimization based upon K-means, K-means Batch Bayesian Optimization (KMBBO). KMBBO was tested in a variety of tasks, from common synthetic functions to the tuning of a machine learning algorithm, to a high-dimensional drug discovery problem.  Over these tasks KMBBO displays superior sampling behaviors than other common Bayesian optimization methods, such as LP, Thompson sampling, Constant Liar, and B3O, delivering either optimal or close to optimal behavior in all tasks.  It also delivered this performance more reliably than any other method, consistently showing the smallest standard deviation in results over 100 repetitions across tasks.  This is a very important result since the major utility of Bayesian optimization is when each sample is expensive or difficult to collect, and thus reliability in optimization performance is strongly desirable.  We also propose a modification to KMBBO, CS-KMBBO, for use in high dimensional problems, where the slice sampling in KMBBO adds significant computational overhead.  In this adaptation, the optimal dimensionality is achieved through the use of the TWiST technique on a sampled subset of the problem space.  CS-KMBBO shows better performance than all methods despite operating on a reduced dimensional data set.  Finally, we discuss insights into the performance of KMBBO through the visualization of the batching process for a toy problem, and comparison to the other methods studied within this paper.  
Over a wide variety of tasks, it is inevitable that for any specific task, a particular sampling technique will have optimal performance, but the strong performance of KMBBO over the whole range of tasks and dimensions, makes it a reliable choice.   

\section{Acknowledgements}
This work was supported by the STFC Hartree Centre’s Innovation Return on Research programme, funded by the Department for Business, Energy \& Industrial Strategy.

\small
\bibliographystyle{aaai}
\bibliography{library}

\end{document}